\documentclass[11pt]{article}

\usepackage[a4paper,margin=1in]{geometry}
\usepackage{newtxtext,newtxmath}

\usepackage{graphicx}
\usepackage{booktabs}
\usepackage{multirow}
\usepackage{caption}
\usepackage{subcaption}   
\usepackage{placeins}     
\usepackage{url}

\usepackage{algorithm}
\usepackage[noend]{algpseudocode}

\usepackage[title]{appendix}

\newcommand{\shorttitle}{Assessing word embeddings after SCM-based bias mitigation}

\makeatletter
\renewcommand{\@oddhead}{\hfil \shorttitle \hfil}
\renewcommand{\@evenhead}{\hfil \shorttitle \hfil}
\makeatother

\usepackage[square,numbers,sort&compress]{natbib}
\usepackage[hidelinks]{hyperref}

\begin{document}

\begin{center}
{\LARGE Assessing the quality and coherence of word embeddings after SCM-based intersectional bias mitigation\par}

{\large
Eren Kocadag$^{1}$,\,
Seyed Sahand Mohammadi Ziabari$^{1,2*}$,\,
Ali Mohammed Mansoor Alsahag$^{1}$
\par}
\vspace{6pt}

{\small
$^{1}$Informatics Institute, University of Amsterdam, 1098 XH Amsterdam, The Netherlands\\
$^{2}$Department of Computer Science and Technology, SUNY Empire State University, Saratoga Springs, NY 12866, USA
\par}
\vspace{6pt}

{\small
E-mail: \href{mailto:eren.kocadag@student.uva.nl}{eren.kocadag@student.uva.nl};
\href{mailto:a.m.m.alsahag@uva.nl}{a.m.m.alsahag@uva.nl};
\href{mailto:sahand.ziabari@sunyempire.edu}{sahand.ziabari@sunyempire.edu}
\par}

{\small
\par}
\end{center}

\vspace{12pt}

\noindent\textbf{Abstract}\quad
Static word embeddings often absorb social biases from the text they learn from, and those biases can quietly shape downstream systems. Prior work that uses the Stereotype Content Model (SCM) has focused mostly on single-group bias along warmth and competence. We broaden that lens to intersectional bias by building compound representations for pairs of social identities through summation or concatenation, and by applying three debiasing strategies: Subtraction, Linear Projection, and Partial Projection. We study three widely used embedding families (Word2Vec, GloVe, and ConceptNet Numberbatch) and assess them with two complementary views of utility: whether local neighborhoods remain coherent and whether analogy behavior is preserved. Across models, SCM-based mitigation carries over well to the intersectional case and largely keeps the overall semantic landscape intact. The main cost is a familiar trade off: methods that most tightly preserve geometry tend to be more cautious about analogy behavior, while more assertive projections can improve analogies at the expense of strict neighborhood stability. Partial Projection is reliably conservative and keeps representations steady; Linear Projection can be more assertive; Subtraction is a simple baseline that remains competitive. The choice between summation and concatenation depends on the embedding family and the application goal.

Together, these findings suggest that intersectional debiasing with SCM is practical in static embeddings, and they offer guidance for selecting aggregation and debiasing settings when balancing stability against analogy performance.
\vspace{6pt}


\section{Introduction}

\label{sec:introduction}

Static word embeddings are a core building block in natural language processing, yet they encode social regularities from their source corpora that can surface as unwanted biases in downstream systems \cite{Papakyriakopoulos2020}. A growing literature has shown how such biases align with human stereotypes and can skew similarity judgments and analogy behavior \cite{Bolukbasi2016, Manzini2019}. To reason about these regularities in a principled way, recent work has adopted the Stereotype Content Model (SCM), a framework from social psychology that maps group perceptions along two interpretable axes, warmth and competence \cite{Fiske2018}. When projected into this two-dimensional space, embeddings expose structured bias that is amenable to targeted attenuation using linear operators \cite{Dev2019,Omrani2023}. Recent work has also extended SCM-guided debiasing to contextual embeddings and to downstream task settings, showing that warmth--competence mitigation can transfer beyond static spaces \cite{Garbat2026WarmthCompetence,Zhu2025TaskAdaptiveSCM}.

Most prior SCM-guided approaches focus on single identities, for example gender or race considered in isolation. In practice, however, people are perceived through intersecting identities, and interactions between social categories can amplify or reshape bias. Intersectional representation therefore presents both a conceptual and a technical challenge for debiasing methods designed around single axes. Extending SCM-based mitigation to the intersectional case requires compound representations that combine identity signals while preserving useful semantics for downstream use.

This paper addresses that gap. We introduce a simple and reproducible pipeline that constructs compound representations for pairs of social identities, fits SCM warmth and competence directions from curated antonym pairs, and applies three standard attenuation operators. We work with three widely used embedding families, which enables model-agnostic conclusions. We evaluate preservation of local structure and analogy behavior using established metrics that measure rank stability and analogy completion.

Concretely, we make three contributions. First, we operationalize intersectional debiasing by building compound embeddings through two constructions, summation and concatenation, and by projecting them into an SCM subspace derived with principal components from carefully selected word-pair differences \cite{Nicolas2021, Omrani2023}. Second, we compare three mitigation operators that trade bias attenuation for utility retention to different degrees, namely Subtraction, Linear Projection, and Partial Projection \cite{Dev2019}. Third, we provide a controlled empirical study across Word2Vec, GloVe, and ConceptNet Numberbatch embeddings \cite{Mikolov2013, Pennington2014, Speer2017}, with light adaptation on WikiText-103 to harmonize vocabularies \cite{Merity2018}, and we quantify effects with the Embedding Coherence Test and the Embedding Quality Test \cite{Dev2019}.

We show how SCM-based mitigation can be extended to intersectional identities without collapsing neighborhoods. We explicitly compare the quality and coherence of compound representations against their single-identity counterparts after debiasing. We also characterize how behavior varies across embedding families and across the two compound constructions, which offers practical guidance for method selection. While we adopt SCM for its interpretability and its prior success in single-identity settings, we acknowledge its cultural sensitivity and use standardized resources to increase transparency \cite{Friehs2022, Gupta2021}. The result is a compact extension of SCM-guided debiasing that is easy to reproduce and that integrates smoothly with existing evaluation practice.

\section{Related Work}
\label{sec:related_work}


As stated in section \ref{sec:introduction}, prior SCM-based bias mitigation efforts focus on individual biases, rather than compound biases, leaving a research gap in addressing compound biases in word embeddings. This section provides an overview of existing research, starting with an introduction to word embeddings and ending with the lack of a standard evaluation metric of intersectional bias in static word embeddings.

Neural-based word embeddings, first introduced by Bengio et al. in 2003 \cite{Bengio2003}, are vector representations of words that preserve semantic relationships. These embeddings capture relationships between words through the contexts in which they are used, enabling machines to understand the semantic similarity of words. This led to the development of more efficient word embeddings, such as Word2Vec \cite{Mikolov2013}, GloVe \cite{Pennington2014}, and Conceptnet Numberbatch \cite{Speer2017}. Although current research often focuses on contextual word embeddings, static word embeddings, as used in this work, are still relevant in research and societal applications. Static word embeddings are computationally less expensive than the contextual variant, and many tasks in NLP still use them \cite{Gupta2021}.

Word embeddings contain biases that impact machine learning models trained on them. These biases may persist even when the data used is supposed to be neutral (e.g., text from Wikipedia). Consequently, societal biases may impact the fairness of models, such as classification or regression models, leading to undesirable decision-making \cite{Papakyriakopoulos2020}. One example of bias in word embeddings is associating neutral words more with male-related words, rather than female-related words. This may exclude female perspectives in NLP applications and reinforce existing gender stereotypes \cite{Petreski2023}. Therefore, it is crucial to minimize the bias present in word embeddings, whether that is gender bias or any other type of bias.

Prior bias mitigation methods address single-dimensional biases in word embeddings. Hard debiasing is one of these methods, specifically designed to mitigate gender bias, by adjusting gender-neutral words to remove gender associations. It also ensures that the embeddings of gender-related word pairs (e.g., father - mother) are equally distant from neutral words \cite{Bolukbasi2016}. However, bias in word embeddings is not limited to gender bias. Social groups may not always be binary but also could contain more than two categories (e.g., race, religion, or occupation). Manzini et al. extend the Hard Debiasing approach to be capable of addressing multi-class biases, while following the same core method \cite{Manzini2019}. Dev and Phillips propose debiasing approaches considered simpler than hard debiasing, while also not relying on manual word-pairing or crowd-sourced labels. The simpler approaches showcased a better preservation of the meaning of the word embeddings after debiasing. Two debiasing methods are introduced: Linear Projection and Partial Projection. Linear projection reduces bias in word embeddings by projecting neutral word embeddings orthogonal to the axis encoding bias. This ensures that the embeddings no longer encode biases, while maintaining their semantic meaning. Partial Projection varies from Linear Projection by controlling how orthogonal the neutral word embeddings need to be to the bias axis. This enables a trade-off between bias reduction and meaning retention \cite{Dev2019}. Although prior bias reduction efforts reduce bias within specific social groups, they are not social-group-agnostic, as they require defining social categories in advance and can only be debiased one at a time. 

To allow bias mitigation efforts to be social-group-agnostic, the Stereotype Content Model (SCM) can be used. The SCM is a social psychology framework that structures stereotypes along two axes: warmth and competence. These axes refer to the perception of friendliness (warmth) and capability (competence) of social groups \cite{Fiske2018}. This enables the SCM to be used as part of a social-group-agnostic approach by representing bias along the same two axes regardless of which social group is examined \cite{Omrani2023}. Omrani et al. build upon the SCM framework by introducing SCM-based debiasing for static word embeddings. They construct word pairs that represent either of the two dimensions (e.g., "sociable" - "unsociable" for warmth or "able" - "unable" for competence). Then, the debiasing methods from Dev and Phillips are used to reduce bias in the embeddings without losing too much meaning of the words from the embeddings \cite{Dev2019}. SCM-based debiasing is effective in reducing bias, while being social-group-agnostic \cite{Omrani2023}. Despite the use of the SCM in this study, it is important to consider that it may introduce its own biases. The SCM attempts to categorize stereotypes into the warmth and competence dimensions universally but does not account for cultural differences in moral judgment. In practice, societies with collectivist values perceive social behavior differently from individualistic cultures, particularly in their approach to social control and the condemning of undesirable behavior \cite{ChenXia2023}. Gonen and Goldberg demonstrate that debiasing methods fail to remove bias entirely, but rather hide it in clusters. The bias is not limited to one axis, but distributed throughout the embedding subspace. This highlights the importance of addressing intersectional biases, which the previously mentioned debiasing approaches have not done \cite{Gonen2019}. More recently, SCM-guided debiasing has been explored beyond static embeddings. Garbat et al.\ investigate warmth--competence debiasing in contextual embedding spaces, providing evidence that SCM directions can be operationalized under contextualization \cite{Garbat2026WarmthCompetence}. Zhu et al.\ further connect SCM debiasing to task behavior via task-adaptive mitigation for sentiment analysis, which motivates our decision to report both structure preservation and analogy-based utility \cite{Zhu2025TaskAdaptiveSCM}.

At the time of writing, there is no commonly agreed upon evaluation metric for intersectional bias in static word embeddings. However, there are previous efforts to create suitable metrics for the evaluation of intersectional biases in word embeddings. Ghai et al. created a visual tool to detect and examine intersectional biases in static word embeddings. In this study, bias was quantified through the Relative Norm Difference (RND) metric, which quantifies how strong the association is between a word and a subgroup compared to another subgroup based on cosine distance \cite{Ghai2021}. Lepori applies Representational Similarity Analysis (RSA) to detect intersectional biases in both static and contextual word embeddings. This method uses Spearman's correlation to compare the pairwise distances between word embeddings to predefined hypothesis models, to quantify how strongly they align. The hypothesis models contain expected biases for word embeddings based on known stereotypes in the real world \cite{Lepori2020}.

Although prior work has addressed bias in word embeddings, no dedicated SCM-based approach has been developed to mitigate intersectional bias in particular. This study proposes an approach to tackle this challenge and fill this research gap.

\section{Data and Problem Setting}

We study intersectional bias mitigation in static word embeddings using publicly available resources and well-defined attribute vocabularies. The task is to construct compound representations for pairs of social identities, map them into a two-dimensional Stereotype Content Model (SCM) subspace (warmth and competence), and apply debiasing while preserving semantic structure measured by ECT and EQT.

\begin{center}
\resizebox{\linewidth}{!}{%
\begin{tabular}{|l|l|l|l|}
\hline
\textbf{Resource} & \textbf{Purpose} & \textbf{Configuration} & \textbf{Notes} \\
\hline
Word2Vec pretrained & Base static embeddings & 300-d vectors & Standard distribution \cite{Mikolov2013} \\
GloVe pretrained & Base static embeddings & 300-d vectors & Common Crawl variant \cite{Pennington2014} \\
ConceptNet Numberbatch & Knowledge-enriched embeddings & 300-d vectors & Multilingual graph-based \cite{Speer2017} \\
WikiText-103 & Light domain adaptation & Unlabeled text & Used only for fine-tuning \cite{Merity2018} \\
SCM word-pair lists & Build warmth/competence axes & Top 15 pairs per axis & From prior work \cite{Nicolas2021,Omrani2023} \\
Group term lists & Identity vocabularies & Gender and race sets & Adapted from \cite{Choenni2021} \\
\hline
\end{tabular}}
\captionof{table}{Corpora and models used in this study. Embedding dimensionality is 300 for all pretrained models.}
\label{tab:data-overview}
\end{center}

\noindent\textit{Identity sets and compound space.}
Let $G = \{g_1,\dots,g_{|G|}\}$ be gender-related terms and $R = \{r_1,\dots,r_{|R|}\}$ race-related terms that are present in the embedding vocabulary. The intersectional set is the Cartesian product
\[
S \;=\; G \times R \;=\; \{(g,r) : g \in G,\; r \in R\},
\]
so the number of compound identities is $|S| = |G| \cdot |R|$. For example, if $|G| = 40$ and $|R| = 40$, then $|S| = 1600$ compound identities.

\noindent\textit{Compound representations.}
Let $e(\cdot) \in \mathbb{R}^{d}$ denote the pretrained vector for a token. For each $(g,r) \in S$ we consider two constructions:
\[
\text{Summation:}\;\; e_{\Sigma}(g,r) \;=\; e(g) + e(r) \in \mathbb{R}^{d}, \qquad
\text{Concatenation:}\;\; e_{\|}(g,r) \;=\; [\,e(g);\; e(r)\,] \in \mathbb{R}^{2d}.
\]
Summation keeps dimensionality fixed and blends signals linearly. Concatenation preserves all coordinates at the cost of doubling dimensionality.

\noindent\textit{SCM bias subspace.}
Warmth and competence are constructed from curated antonym pairs. Let $\mathcal{W}$ and $\mathcal{C}$ be the top-$k$ word-pair sets (here $k=15$ per axis). For each pair $(a^+,a^-)\in\mathcal{W}$ we form a difference vector $v = e(a^+) - e(a^-)$; stacking these gives matrix $V_{\mathcal{W}} \in \mathbb{R}^{d \times k}$. The warmth direction $u_{\mathcal{W}}$ is the first principal component of $V_{\mathcal{W}}$. An analogous procedure yields the competence direction $u_{\mathcal{C}}$. These two directions span the SCM subspace used for projection and debiasing \cite{Omrani2023,Dev2019}.

\noindent\textit{Coverage and token hygiene.}
Let $\mathcal{V}_m$ be the vocabulary of model $m$. Coverage for a term set $T$ under $m$ is
\[
\text{cov}_m(T) \;=\; \frac{|\,T \cap \mathcal{V}_m\,|}{|T|}.
\]
Before experiments, out-of-vocabulary items were replaced by close synonyms or removed to keep $\text{cov}_m(T)$ high for all models (procedure detailed in Method). This ensures consistency across Word2Vec, GloVe, and Numberbatch.

\noindent\textit{Problem statement.}
Given the compound set $S$ and SCM directions $\{u_{\mathcal{W}}, u_{\mathcal{C}}\}$, learn a debiased representation $\tilde{e}(\cdot)$ from a chosen operator $\mathcal{D} \in \{\text{Subtraction, Linear Projection, Partial Projection}\}$ such that:
\[
\text{(i) Bias reduction:}\;\; \forall (g,r)\in S,\; \text{projections onto } u_{\mathcal{W}}, u_{\mathcal{C}} \text{ are attenuated};
\]
\[
\text{(ii) Utility preservation:}\;\; \text{relative neighborhoods and analogy behavior are retained.}
\]
Utility is quantified by ECT (rank stability) and EQT (analogy completion), reported in Section~\ref{sec:results}. The data choices above fix the problem setting while enabling controlled comparisons across models and compound constructions.

\section{Method}
\label{sec:Method}



This section presents the experimental setup of this study to ensure reproducibility. The experimental setup closely follows the approach of Omrani et al \cite{Omrani2023}. First, the data and word embeddings used are described. Next, the bias subspace based on the SCM model is explained. Then, the bias mitigation methods applied to reduce bias in the word embeddings are detailed. Finally, an evaluation setup details the metrics used to measure bias reduction.

In Figure \ref{fig:flowchart}, a flowchart of the general approach of the methodology of this study is displayed.

\begin{center}
  \includegraphics[width=0.8\linewidth]{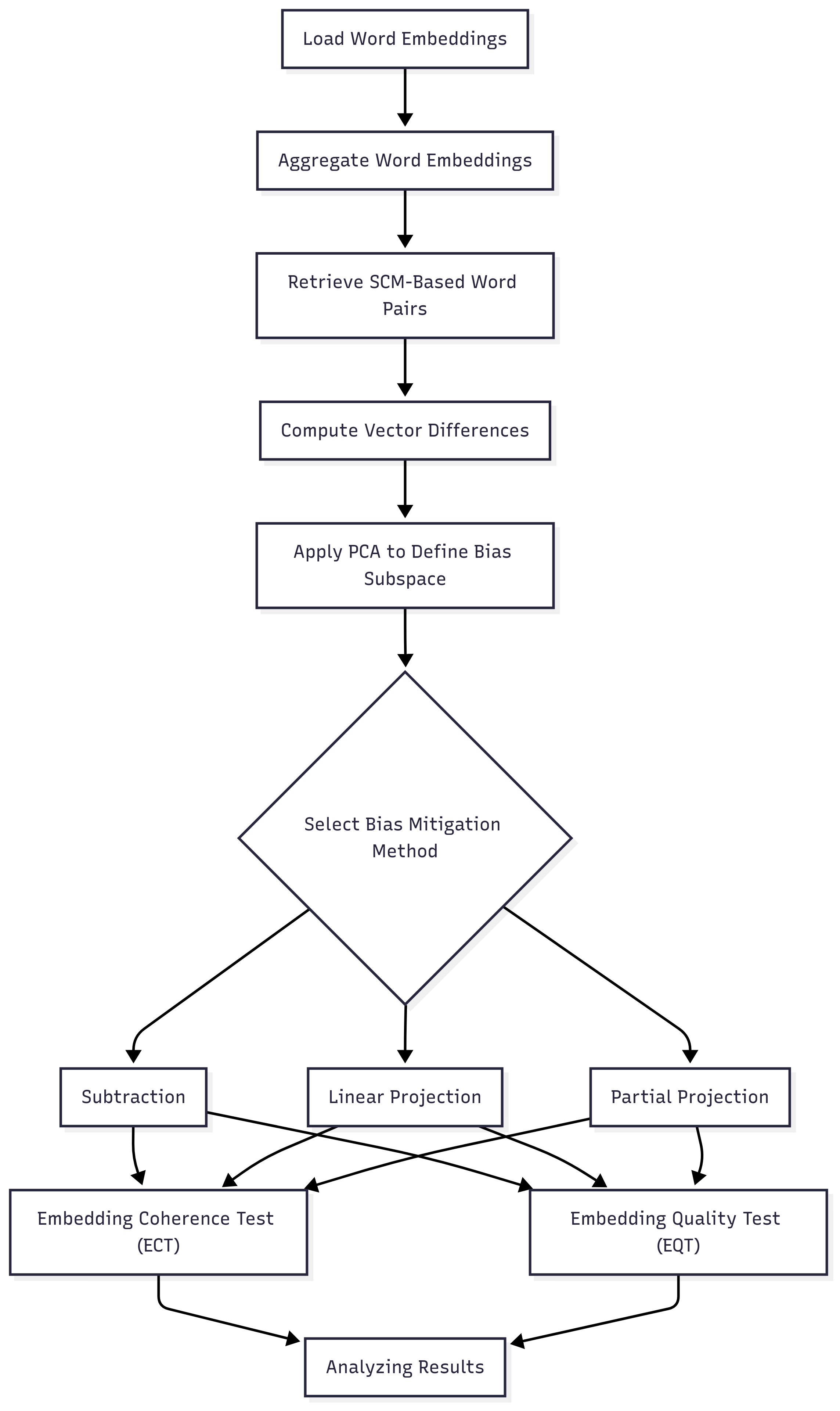}
  \captionof{figure}{An overview of the methodology for bias mitigation and for evaluating the coherence and quality of the word embeddings.}
  \label{fig:flowchart}
\end{center}

\subsection{Word embeddings}
\label{subsec: word embeddings}
To thoroughly investigate bias in static word embeddings, this study uses three pre-trained static word embeddings: Word2Vec, GloVe, and Conceptnet Numberbatch. This selection includes both predictive and count-based approaches, providing a well-rounded selection. Word2Vec utilizes the local contexts of words to learn word embeddings based on predicted co-occurrences \cite{Mikolov2013}.  
GloVe uses global co-occurrences to construct word embeddings reflecting the entire corpus \cite{Pennington2014}. Conceptnet Numberbatch goes beyond distributional semantics by constructing the embeddings from a knowledge graph, enabling the embeddings to capture ontological information \cite{Speer2017}. By using a variety of models, this study ensures a diverse basis for bias evaluation.

In line with Omrani et al., the embeddings are loaded with the Gensim library and fine-tuned on the Wikitext-103 corpus \cite{Omrani2023, Merity2018}. This approach maintains most of the word embeddings as they were initially trained, but adapts to the general language of the Wikipedia text. There was no domain-specific fine-tuning, as this study aims to investigate SCM-based bias mitigation for generic word embeddings not catered to one social category.

Static word embeddings are trained to have embeddings for individual words, meaning that compound identities are not explicitly captured. Previous studies have shown that gender and race bias have been the most investigated, motivating this work to use the same groups \cite{Ghai2021}.
Considering that bias mitigation in word embeddings is performed in this study, the method for constructing compound word embeddings should minimize information loss and preserve the semantic relationships of the compound word embeddings. This study uses two aggregation methods to create compound word embeddings from two individual word embeddings:
\begin{itemize}
    \item \textbf{Concatenation:} The word embeddings of two words are concatenated to construct a single vector. This ensures that no semantic information is lost, but doubles the dimensionality of the word embeddings used. \\
    Example: $[1,\ 2] \| [3,\ 4] = [1,\ 2,\ 3,\ 4]$
    \item \textbf{Summing:} The word embeddings are summed into a single vector through their elementwise sum. This method retains the original embedding dimensionality and is intuitive from a linear algebra perspective. \\
    Example: $[1,\ 2] + [3,\ 4] = [4,\ 6]$
\end{itemize}
 
 These aggregation methods are applied consistently across Word2Vec, GloVe, and Conceptnet Numberbatch, enabling a comparative analysis of the aggregation methods and the word embedding models used.

Other aggregation methods, such as averaging or weighted combinations, were omitted from this study. Averaging was excluded to avoid information loss from compressing the embeddings by dividing them by two. Weighted combinations introduce additional complexity in aggregation by adding the need to determine how strongly one social category weighs over the other in aggregated embeddings. This may reduce the interpretability of the results and make comparisons between other social groups more difficult in future studies. In contrast, concatenation retains all semantic information from both embeddings, while summing retains the original dimensionality and finds a linear algebraic middle ground between two terms. These methods provide a balance between limiting information loss and maintaining generalizability.
 \subsection{Word pairs}
 \label{subsec:wordpairs}
To construct the bias subspace and have relevant words for debiasing, Omrani et al. use word pairs from prior work that represent the axes of the SCM framework \cite{Omrani2023,Nicolas2021}. These word pairs can be used for one-dimensional bias mitigation and extending them for compound embeddings may provide new challenges.

As mentioned in section \ref{subsec: word embeddings}, this study focuses on compound embeddings that represent gender and race. The previously constructed lists of word pairs for race only contain names and, for gender, occasionally contain pronouns. The race-related terms are also limited in quantity. When compound word embeddings are created using combinations from these existing word pairs, the combinations may lack real semantic meaning and result in insufficient samples.

To address these limitations, this study builds on term lists by Choenni et al., which particularly contain a more diverse and informative list of words pertaining to race \cite{Choenni2021}. These terms are adapted to fit the objectives of this work. Since there is a list of gender-related terms and one of race-related terms, these need to be concatenated and stored in the embedding model as described in \ref{subsec: word embeddings}. The terms used in the Omrani et al. are shown in Figure \ref{fig:ogterms}, while the terms inspired by Choenni et al. are displayed in Figure \ref{fig:newterms}.

\begin{center}
  \includegraphics[width=0.45\textwidth]{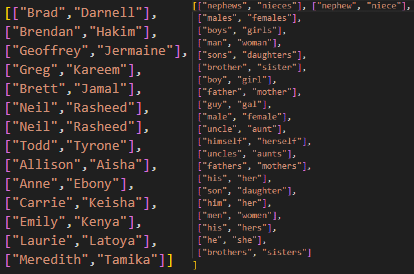}
  \captionof{figure}{The original terms used by Omrani et al. \cite{Omrani2023}, with the race-related terms on the left side and the gender-related terms on the right.}
  \label{fig:ogterms}
\end{center}

\begin{center}
  \includegraphics[width=0.45\textwidth]{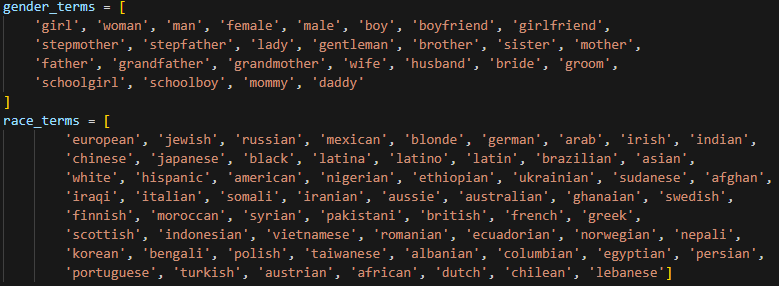}
  \captionof{figure}{The terms used for bias mitigation in this study inspired by Choenni et al. \cite{Choenni2021}. The words in the first list are related to gender, while the words in the second list are related to race.}
  \label{fig:newterms}
\end{center}

 Constructing contrastive word pairs for compound word embeddings is considerably more challenging than for single-group embeddings, as the combinatorial nature of the aggregation methods leads to a high number of compound terms. With more than 1600 compound word embeddings, it is impractical and time-intensive to construct contrastive word pairs. To address this, random word pairing was applied by shuffling a list of aggregated embeddings and pairing every embedding with another. While this does not ensure meaningful semantic relations between resulting pairs, it enables a scalable and unbiased experimental setup.

To mitigate the risk of unrepresentative pairings impacting the results, the random pairing process is repeated three times for each combination of word embedding model and aggregation method. This means that each configuration of the experiment is performed three times to aid the reliability and robustness of the findings in this study. Due to time constraints, no further steps were taken to enforce or check meaningful semantic relations between word pairs.

 A noteworthy decision may be not to use dynamic word pair selection. This decision was made with computational constraints in mind, as implementing such a method would increase the already substantial runtime of the experimental setup. Additionally, it would provide a design challenge, as this study aims to address intersectional biases. It would be difficult to define what compound word pairs are contrastive in a dynamic setting.
\subsection{Defining the bias subspace}
\label{subsec:biassubspace}
This study constructs the bias subspace using the SCM framework mentioned in section \ref{sec:introduction}, which defines warmth and competence as the bias dimensions. This bias subspace will then be used to apply bias mitigation methods on.

First, word pairs as mentioned in section \ref{subsec:wordpairs} are loaded in from corresponding JSON files. Examples of such word pairs could be "Genuine" and "Fake" for warmth or "Able" and "Unable" for competence. Then, following the approach of Omrani et al., the top 15 word pairs for each dimension are sampled to ensure that the number of word pairs is not too large \cite{Omrani2023}. Whether or not embeddings exist for each word in the downsampled pairs in all three used word embeddings must also be checked. If a word is missing, it may be replaced with a similar term, or the word pair may be discarded for an alternative. Similarity could be evaluated through cosine similarity.

The bias subspace is constructed so that the compound groups can be projected onto it afterward. First, the word embeddings from each word of the top 15 word pairs of both dimensions are extracted from Word2Vec, GloVe, and Conceptnet Numberbatch. This is done to prevent the number of word pairs from becoming too large, which may introduce redundancy. Then for each word pair of each pre-trained word embedding, the difference between the vector representations of the words is calculated by subtracting the vector representation of the negative word (e.g., "Fake") from the vector representation of the positive word (e.g., "Genuine"). These differences are then stacked into two matrices, one for warmth and one for competence. Next, \textit{Principal Component Analysis} (PCA) is applied separately to these two vectors to find their respective principal components, defining the bias subspace. The first principal component is selected for both warmth and competence, which serves as a foundation for debiasing that captures the primary direction of bias along its dimension. 
\subsection{Bias mitigation methods}
The bias mitigation methods used in this study are inspired by the work of Omrani et al \cite{Omrani2023}. They employ Subtraction, Hard Debiasing, Linear Projection, and Partial Projection. Hard Debiasing was excluded from this study, as it specifically addresses gender bias and does not apply to intersectional bias. However, it is possible to implement Subtraction, Linear Projection, and Partial Projection, as described below.
\begin{itemize}
    
    \item \textbf{Linear Projection} removes bias by projecting word embeddings orthogonally to the main bias direction lying in the bias subspace, denoted as \(v_B\) \cite{Dev2019}.  For all word embeddings \(w\), the debiased word embedding \(w'\) is defined as: $$ w' \mathrel{\mathop:}= w - \pi_B$$
    Here \(\pi_B\) is the projection of \(w\) onto \(v_B\).

    Linear projection provides strict debiasing, and serves as a baseline for partial projection.
    \item \textbf{Partial Projection} is an extension of linear projection \cite{Dev2019}. It introduces the possibility to control the extent to which embeddings are projected away from the bias direction. This is done by applying a scaling function that balances bias removal with semantic retention. Therefore, it is more conservative than linear projection.. Partial projection is formalized as seen below.
    $$ w' = \mu + r(w) + \beta \cdot f(\|r(w)\|) \cdot v_B $$
    Where:
    \begin{itemize}
        \item  \( \mu \) is the mean of the word embeddings defining the bias subspace.
        \item \(r(w) = w - \text{Proj}_{v_B}(w)\), is the component of \(w\) orthogonal to the bias direction.
        \item \(\beta\) is a scalar controlling the extent of the bias removal.
        \item \(f(.)\) is a smoothing function that contributes to reducing unintended bias and preserving definitional bias.

    \end{itemize}
    \item \textbf{Subtraction} is a baseline proposed by Dev and Phillips \cite{Dev2019}, which subtracts the bias direction  \(v_B\) from all word embeddings. This is approach is computationally simple, but may be less precise than the other methods. Subtraction could be formalized by stating that for all word embeddings \(w\), the debiased word embedding \(w'\) can be defined as: $$ w' \mathrel{\mathop:}= w - v_B$$
\end{itemize}
\subsection{Evaluation}
To evaluate the coherence and quality of the word embeddings, two evaluation metrics are applied on the compound word embedding after debiasing. The following evaluation metrics are considered. 
\begin{itemize}
    \item \textbf{Embedding Coherence Test (ECT)}:
Suppose \(A\) is a set of word pairs: \(\{(a_1^1, a_1^2, ...), (a_2^1, a_2^2, ...)\}\), where \(a_i^j\) represents the j-th word of the i-th subgroup of an attribute, with each subgroup being related to warmth or competence. Warmth-related terms could be ("Friendly" - "Cold"), while competence-related terms could be ("Intelligent" - "Unintelligent"). Then, also suppose that \(P\) is a set of compound social identities \(\{p_1, p_2, ...\}\)\\
    ECT then computes the Spearman rank correlation between the ranking of cosine similarities between \(P\) and \(A\) before and after the debiasing of the word embeddings of \(P\). A high score indicates that social groups maintain relative similarity to the terms of warmth and competence after debiasing, thus retaining their semantic relationships.
    \item \textbf{Embedding Quality Test (EQT)}:
    The EQT metric measures whether word embeddings retain their ability to correctly complete analogies after debiasing. Like ECT, EQT uses a set of word pairs \(A\) and a set of compound social identities \(P\).

    Each word pair consists of a positive term and a negative term, which function as the basis for an analogy. The analogy contains a starting term and an expected completion, which the word embedding models complete. EQT tests whether the word embeddings correctly identify the expected term to complete the analogy.

    A correct and unbiased response would be the expected term, its plural form, or a synonym. Then the proportion of correct responses is the EQT score. A higher EQT score indicates that word relationships are maintained after debiasing, which means that the semantics of the word embeddings are preserved.
\end{itemize}
The results of this study are evaluated against those presented by Omrani et al. for the mitigation of single-dimensional bias. However, because this study debiases different terms than Omrani et al., a direct comparison may not be valid \cite{Omrani2023}. To address this, a baseline is created by generating random word pairs individually for each category of terms (race and gender). This follows the procedure described in section \ref{subsec:wordpairs}. The same experimental setup outlined in this methodology is then applied separately to both sets of word pairs. These single-dimensional debiasing results serve as a more appropriate baseline for evaluating the results of multi-dimensional debiasing.

To assess the significance of the results, each run of the experimental setup is evaluated using p-values. This is done for each combination of a word embedding model and an aggregation method across three repetitions.

\section{Data and Problem Setting}
\label{sec:Data and Problem Setting}
We study intersectional bias mitigation in static word embeddings using publicly available resources and well-defined attribute vocabularies. The task is to construct compound representations for pairs of social identities, map them into a two-dimensional Stereotype Content Model (SCM) subspace (warmth and competence), and apply debiasing while preserving semantic structure measured by ECT and EQT.

\begin{center}
\resizebox{\linewidth}{!}{%
\begin{tabular}{|l|l|l|l|}
\hline
\textbf{Resource} & \textbf{Purpose} & \textbf{Configuration} & \textbf{Notes} \\
\hline
Word2Vec pretrained & Base static embeddings & 300-d vectors & Standard distribution \cite{Mikolov2013} \\
GloVe pretrained & Base static embeddings & 300-d vectors & Common Crawl variant \cite{Pennington2014} \\
ConceptNet Numberbatch & Knowledge-enriched embeddings & 300-d vectors & Multilingual graph-based \cite{Speer2017} \\
WikiText-103 & Light domain adaptation & Unlabeled text & Used only for fine-tuning \cite{Merity2018} \\
SCM word-pair lists & Build warmth/competence axes & Top 15 pairs per axis & From prior work \cite{Nicolas2021,Omrani2023} \\
Group term lists & Identity vocabularies & Gender and race sets & Adapted from \cite{Choenni2021} \\
\hline
\end{tabular}}
\captionof{table}{Corpora and models used in this study. Embedding dimensionality is 300 for all pretrained models.}
\label{tab:data-overview}
\end{center}

\noindent\textit{Identity sets and compound space.}
Let $G = \{g_1,\dots,g_{|G|}\}$ be gender-related terms and $R = \{r_1,\dots,r_{|R|}\}$ race-related terms that are present in the embedding vocabulary. The intersectional set is the Cartesian product
\[
S \;=\; G \times R \;=\; \{(g,r) : g \in G,\; r \in R\},
\]
so the number of compound identities is $|S| = |G| \cdot |R|$. For example, if $|G| = 40$ and $|R| = 40$, then $|S| = 1600$ compound identities.

\noindent\textit{Compound representations.}
Let $e(\cdot) \in \mathbb{R}^{d}$ denote the pretrained vector for a token. For each $(g,r) \in S$ we consider two constructions:
\[
\text{Summation:}\;\; e_{\Sigma}(g,r) \;=\; e(g) + e(r) \in \mathbb{R}^{d}, \qquad
\text{Concatenation:}\;\; e_{\|}(g,r) \;=\; [\,e(g);\; e(r)\,] \in \mathbb{R}^{2d}.
\]
Summation keeps dimensionality fixed and blends signals linearly. Concatenation preserves all coordinates at the cost of doubling dimensionality.

\noindent\textit{SCM bias subspace.}
Warmth and competence are constructed from curated antonym pairs. Let $\mathcal{W}$ and $\mathcal{C}$ be the top-$k$ word-pair sets (here $k=15$ per axis). For each pair $(a^+,a^-)\in\mathcal{W}$ we form a difference vector $v = e(a^+) - e(a^-)$; stacking these gives matrix $V_{\mathcal{W}} \in \mathbb{R}^{d \times k}$. The warmth direction $u_{\mathcal{W}}$ is the first principal component of $V_{\mathcal{W}}$. An analogous procedure yields the competence direction $u_{\mathcal{C}}$. These two directions span the SCM subspace used for projection and debiasing \cite{Omrani2023,Dev2019}.

\noindent\textit{Coverage and token hygiene.}
Let $\mathcal{V}_m$ be the vocabulary of model $m$. Coverage for a term set $T$ under $m$ is
\[
\text{cov}_m(T) \;=\; \frac{|\,T \cap \mathcal{V}_m\,|}{|T|}.
\]
Before experiments, out-of-vocabulary items were replaced by close synonyms or removed to keep $\text{cov}_m(T)$ high for all models (procedure detailed in Method). This ensures consistency across Word2Vec, GloVe, and Numberbatch.

\noindent\textit{Problem statement.}
Given the compound set $S$ and SCM directions $\{u_{\mathcal{W}}, u_{\mathcal{C}}\}$, learn a debiased representation $\tilde{e}(\cdot)$ from a chosen operator $\mathcal{D} \in \{\text{Subtraction, Linear Projection, Partial Projection}\}$ such that:
\[
\text{(i) Bias reduction:}\;\; \forall (g,r)\in S,\; \text{projections onto } u_{\mathcal{W}}, u_{\mathcal{C}} \text{ are attenuated};
\]
\[
\text{(ii) Utility preservation:}\;\; \text{relative neighborhoods and analogy behavior are retained.}
\]
Utility is quantified by ECT (rank stability) and EQT (analogy completion), reported in Section~\ref{sec:results}. The data choices above fix the problem setting while enabling controlled comparisons across models and compound constructions.

\section{Results}
\label{sec:results}
The results of the experiments using the Word2Vec, GloVe, and ConceptNet Numberbatch word embedding models are presented in Tables~\ref{tab:Word2Vec}, \ref{tab:GloVe}, and \ref{tab:C}. Table~\ref{tab:aggmethods} summarizes the results across aggregation methods. All values are averaged over three independent runs of the experimental setup. For every configuration and repetition, the observed differences are statistically significant, with p-values below 0.05. This section provides a comparative analysis of aggregation strategies, debiasing targets, and embedding models.

Across Tables~\ref{tab:Word2Vec}, \ref{tab:GloVe}, \ref{tab:C}, and \ref{tab:aggmethods}, both aggregation methods, Summation and Concatenation, lead to broadly comparable outcomes. However, the aggregation strategy yielding the highest Embedding Coherence Test (ECT) and Embedding Quality Test (EQT) scores varies by embedding model. For Word2Vec, concatenation consistently results in higher ECT and EQT scores. In contrast, ConceptNet Numberbatch benefits more from summation, particularly with respect to EQT. GloVe exhibits a more balanced pattern, with only marginal differences between the two aggregation strategies.

A comparison between compound representations and single-dimensional debiasing targets shows that compound word embeddings generally achieve higher ECT scores, indicating stronger preservation of relative similarity structure after debiasing. At the same time, EQT scores tend to be higher for single-dimensional debiasing than for compound representations, suggesting that analogy completion becomes more sensitive as multiple social dimensions are combined. ConceptNet Numberbatch constitutes a partial exception, with EQT scores that remain comparable between single-dimensional and compound settings.

\begin{center}
\resizebox{\linewidth}{!}{%
\begin{tabular}{|c|c|c|c|c|}
\hline
 & Race & Gender & Summed & Concatenated\\
\hline
ECT\(_{Sub}\) & 0.9461 \(\pm\) 0.004 & 0.958 \(\pm\) 0.003 & 0.998 \(\pm\) 0& \textbf{0.9989 \(\pm\) 0.001} \\
EQT\(_{Sub}\) & \textbf{0.0569 \(\pm\) 0.005} & 0.0426 \(\pm\) 0.003 & 0.0291 \(\pm\) 0& 0.0335 \(\pm\) 0.001\\
ECT\(_{LP}\) & 0.9364 \(\pm\) 0.004& 0.9581 \(\pm\) 0.003 & 0.9981 \(\pm\) 0& \textbf{0.9989 \(\pm\) 0.001}\\
EQT\(_{LP}\) & \textbf{0.0586 \(\pm\) 0.006} & 0.0446 \(\pm\) 0.004 & 0.0285 \(\pm\) 0& 0.0322 \(\pm\) 0.001\\
ECT\(_{PP}\) & 0.9981 \(\pm\) 0 & 0.964 \(\pm\) 0.022& 0.9993 \(\pm\) 0& \textbf{0.9998 \(\pm\) 0}\\
EQT\(_{PP}\) & \textbf{0.0428 \(\pm\) 0.005} & 0.0332 \(\pm\) 0.003 & 0.0186 \(\pm\) 0& 0.0213 \(\pm\) 0\\
\hline
\end{tabular}
}
\captionof{table}{ECT and EQT for the Word2Vec model for both single-dimensional bias reduction (Race, Gender) and the multi-dimensional bias reduction (Summed, Concatenated). Averages for three runs across different debiasing methods are shown (Sub, LP, PP), with corresponding standard deviations rounded to three decimals.}
\label{tab:Word2Vec}
\end{center}

\begin{center}
\resizebox{\linewidth}{!}{%
\begin{tabular}{|c|c|c|c|c|}
\hline
 & Race & Gender & Summed & Concatenated\\
\hline
ECT\(_{Sub}\) & 0.973 \(\pm\) 0.009& 0.9507 \(\pm\) 0.005& \textbf{0.9987 \(\pm\) 0}& 0.9986 \(\pm\) 0\\
EQT\(_{Sub}\) & 0.1993 \(\pm\) 0.002& \textbf{0.2301 \(\pm\) 0.047} & 0.1217 \(\pm\) 0.002& 0.1222 \(\pm\) 0.001\\
ECT\(_{LP}\) & 0.9554 \(\pm\) 0.015& 0.9399 \(\pm\) 0.008& \textbf{0.9986 \(\pm\) 0}& 0.9985 \(\pm\) 0.001\\
EQT\(_{LP}\) & 0.1875 \(\pm\) 0.002& \textbf{0.2290 \(\pm\) 0.048} & 0.1184 \(\pm\) 0.002& 0.1265 \(\pm\) 0.001\\
ECT\(_{PP}\) & 0.9775 \(\pm\) 0.009& 0.9810 \(\pm\) 0.002& 0.999 \(\pm\) 0& \textbf{0.9995 \(\pm\) 0}\\
EQT\(_{PP}\) & 0.1818 \(\pm\) 0.002& \textbf{0.2513 \(\pm\) 0.047} & 0.0839 \(\pm\) 0.001& 0.0903 \(\pm\) 0.001\\
\hline
\end{tabular}
}
\captionof{table}{ECT and EQT for the GloVe model for both single-dimensional bias reduction (Race, Gender) and the multi-dimensional bias reduction (Summed, Concatenated). Averages for three runs across different debiasing methods are shown (Sub, LP, PP), with corresponding standard deviations rounded to three decimals.}
\label{tab:GloVe}
\end{center}

\begin{center}
\resizebox{\linewidth}{!}{%
\begin{tabular}{|c|c|c|c|c|}
\hline
 & Race & Gender & Summed & Concatenated\\
\hline
ECT\(_{Sub}\) & 0.9268 \(\pm\) 0.011& 0.8903 \(\pm\) 0.013& \textbf{0.9984 \(\pm\) 0}& 0.9982 \(\pm\) 0\\
EQT\(_{Sub}\) & \textbf{0.1306 \(\pm\) 0.004} & 0.1059 \(\pm\) 0.001& 0.1169 \(\pm\) 0.001& 0.1127 \(\pm\) 0.001\\
ECT\(_{LP}\) & 0.9008 \(\pm\) 0.010& 0.8649 \(\pm\) 0.015& \textbf{0.9984 \(\pm\) 0}& 0.9979 \(\pm\) 0\\
EQT\(_{LP}\) & 0.1326 \(\pm\) 0.004& 0.1059 \(\pm\) 0.001& \textbf{0.1431 \(\pm\) 0.001} & 0.1380 \(\pm\) 0\\
ECT\(_{PP}\) & 0.9333 \(\pm\) 0.004& 0.8928 \(\pm\) 0.013& \textbf{0.9987 \(\pm\) 0}& 0.9982 \(\pm\) 0\\
EQT\(_{PP}\) & 0.1259 \(\pm\) 0.006& 0.1066 \(\pm\) 0.002& \textbf{0.1268 \(\pm\) 0.001}& 0.1235 \(\pm\) 0.001\\
\hline
\end{tabular}
}
\captionof{table}{ECT and EQT for the ConceptNet Numberbatch model for both single-dimensional bias reduction (Race, Gender) and the multi-dimensional bias reduction (Summed, Concatenated). Averages for three runs across different debiasing methods are shown (Sub, LP, PP), with corresponding standard deviations rounded to three decimals.}
\label{tab:C}
\end{center}

\begin{center}
\begin{tabular}{|c|c|c|}
\hline
Aggregation Method & Mean ECT & Mean EQT\\
\hline
Summation & 0.9986 & 0.1031\\
Concatenation & 0.9988 & 0.0889\\
\hline
\end{tabular}
\captionof{table}{Mean ECT and EQT scores across all embedding models and debiasing methods.}
\label{tab:aggmethods}
\end{center}

Regarding differences across embedding families, analogy behavior shows the clearest variation. While ECT scores remain consistently high across all models, Word2Vec produces substantially lower EQT scores than GloVe and ConceptNet Numberbatch across most configurations. For single-dimensional debiasing, GloVe yields higher ECT scores than the other embedding models. These differences are reduced in the intersectional setting, where ECT values converge across models. Overall, the results indicate that SCM-based bias mitigation preserves embedding coherence across models, while changes in analogy performance are more strongly influenced by the underlying embedding architecture than by the choice of aggregation strategy.

\section{Ablations and Analysis}
\label{sec:ablations}
This section examines three simple ablations that help interpret the main findings without introducing any new data sources or models. We focus on how the choice of aggregation, the choice of debiasing method, and the choice of target (single group versus intersectional) affect the two evaluation metrics defined in Section~\ref{sec:Method}. All numbers below are aggregated from the Results tables and are intended to be read as directional evidence rather than definitive benchmarks.

\paragraph{Ablation A1: Effect of the aggregation strategy.}
Table~\ref{tab:agg_effect} contrasts summation and concatenation for each embedding model by averaging over the three debiasers. For Word2Vec, concatenation increases ECT a little and also increases EQT slightly. For GloVe, both strategies are nearly indistinguishable on ECT, with concatenation giving a small increase in EQT. For ConceptNet Numberbatch, summation yields marginally higher ECT and also a slightly higher EQT. These patterns suggest that the best aggregation strategy is model dependent, and the differences are small in magnitude.

\begin{center}
\resizebox{\linewidth}{!}{%
\begin{tabular}{|l|c c c|c c c|}
\hline
\multirow{2}{*}{Model} & \multicolumn{3}{c|}{ECT (mean over debiasers)} & \multicolumn{3}{c|}{EQT (mean over debiasers)} \\
 & Summation & Concatenation & $\Delta$ECT & Summation & Concatenation & $\Delta$EQT \\
\hline
Word2Vec & 0.9985 & 0.9992 & {+}0.0007 & 0.0254 & 0.0290 & {+}0.0036 \\
GloVe & 0.9988 & 0.9989 & {+}0.0001 & 0.1080 & 0.1130 & {+}0.0050 \\
ConceptNet & 0.9985 & 0.9981 & $-$0.0004 & 0.1289 & 0.1247 & $-$0.0042 \\
\hline
\end{tabular}
}
\captionof{table}{Aggregation ablation. Mean ECT and EQT for summation and concatenation, averaged over Subtraction, Linear Projection, and Partial Projection. Deltas are Concatenation minus Summation.}
\label{tab:agg_effect}
\end{center}

\paragraph{Ablation A2: Effect of the debiasing method.}
Table~\ref{tab:debiaser_effect} averages each debiaser across both aggregation strategies. Partial Projection consistently delivers the highest ECT and the lowest EQT for Word2Vec and GloVe, indicating a conservative behavior that preserves neighborhood structure while limiting analogy disruptions. For ConceptNet Numberbatch, Partial Projection still leads on ECT, although Linear Projection gives the largest EQT, which reflects a more aggressive removal of bias directions that can sometimes help analogy completions.

\begin{center}
\resizebox{\linewidth}{!}{%
\begin{tabular}{|l|c c c|c c c|}
\hline
\multirow{2}{*}{Model} & \multicolumn{3}{c|}{ECT (avg over aggregations)} & \multicolumn{3}{c|}{EQT (avg over aggregations)} \\
 & Subtraction & Linear Proj. & Partial Proj. & Subtraction & Linear Proj. & Partial Proj. \\
\hline
Word2Vec & 0.9985 & 0.9985 & \textbf{0.9996} & 0.0313 & 0.0304 & \textbf{0.0200} \\
GloVe & 0.9986 & 0.9986 & \textbf{0.9993} & 0.1220 & 0.1225 & \textbf{0.0871} \\
ConceptNet & 0.9983 & 0.9982 & \textbf{0.9985} & 0.1148 & \textbf{0.1406} & 0.1252 \\
\hline
\end{tabular}
}
\captionof{table}{Debiaser ablation. Each cell averages the metric over Summation and Concatenation for the given debiaser. Bold indicates the best value per row.}
\label{tab:debiaser_effect}
\end{center}

\paragraph{Ablation A3: Single group versus intersectional targets.}
Table~\ref{tab:single_vs_intersectional} compares averages across the three debiasers for Race only, Gender only, and the Intersectional setting which averages across both aggregation strategies. Intersectional ECT is very high across all models, which indicates strong preservation of rank order with respect to warmth and competence after debiasing. For Word2Vec and GloVe, EQT declines when moving from single groups to intersectional targets, which is consistent with the intuition that analogy completions become more delicate when two social axes are combined. For ConceptNet Numberbatch, intersectional EQT sits between the Race and Gender single conditions, which points to a different tradeoff surface for knowledge graph based embeddings.

\begin{center}
\resizebox{\linewidth}{!}{%
\begin{tabular}{|l|c c|c c|c c|}
\hline
\multirow{2}{*}{Setting} & \multicolumn{2}{c|}{Word2Vec} & \multicolumn{2}{c|}{GloVe} & \multicolumn{2}{c|}{ConceptNet} \\
 & ECT & EQT & ECT & EQT & ECT & EQT \\
\hline
Race only & 0.9602 & 0.0528 & 0.9686 & 0.1895 & 0.9203 & 0.1297 \\
Gender only & 0.9600 & 0.0401 & 0.9572 & 0.2368 & 0.8827 & 0.1061 \\
Intersectional & \textbf{0.9988} & 0.0272 & \textbf{0.9988} & 0.1105 & \textbf{0.9983} & 0.1268 \\
\hline
\end{tabular}
}
\captionof{table}{Single group versus intersectional ablation. Each entry is an average over the three debiasers. Intersectional values also average over both aggregation strategies.}
\label{tab:single_vs_intersectional}
\end{center}

\paragraph{Interpretation and practical guidance.}
Across all ablations, ECT remains near one in the intersectional setting, which means the debiased embeddings preserve their relative alignment to warmth and competence vocabulary. EQT is more sensitive to design choices. If preserving analogy behavior is the priority, ConceptNet Numberbatch with Linear Projection or Subtraction can yield higher EQT at the cost of slightly lower ECT. If preserving rank structure is the priority, Partial Projection is a safe default for Word2Vec and GloVe. For aggregation, start with summation for ConceptNet Numberbatch and with concatenation for Word2Vec, while expecting very small differences for GloVe. These recommendations follow directly from the compact averages above and are consistent with the qualitative conclusions in Section~\ref{sec:discussion}.

\section{Discussion}
\label{sec:discussion}
Our results show that SCM-guided mitigation scales cleanly to intersectional settings without collapsing local neighborhoods. After debiasing, ECT remains high across models and compound constructions, indicating that relative similarity structure is largely preserved. EQT decreases modestly, which is expected when attenuating targeted variance while evaluating with a brittle analogy probe that penalizes both biased and nonsensical completions equally \cite{Dev2019}.

Model behavior is consistent but not identical. Word2Vec is more sensitive on EQT, likely due to its reliance on local co-occurrence statistics. GloVe maintains strong ECT for single-group settings, and ConceptNet Numberbatch is comparatively robust once compound identities are introduced. These patterns match prior findings summarized in Table~\ref{tab:Omrani} and help in choosing a mitigation operator for a given tolerance to change.

Operator choice offers a practical dial. Partial Projection produces conservative adjustments with very strong ECT. Linear Projection and Subtraction achieve stronger attenuation at a higher cost to analogy behavior. In applications where neighborhood stability is the priority, Partial Projection is a safe default. When stronger bias reduction is required and small utility shifts are acceptable, Linear Projection or Subtraction can be preferable \cite{Dev2019}.

\begin{center}
  \includegraphics[width=\linewidth]{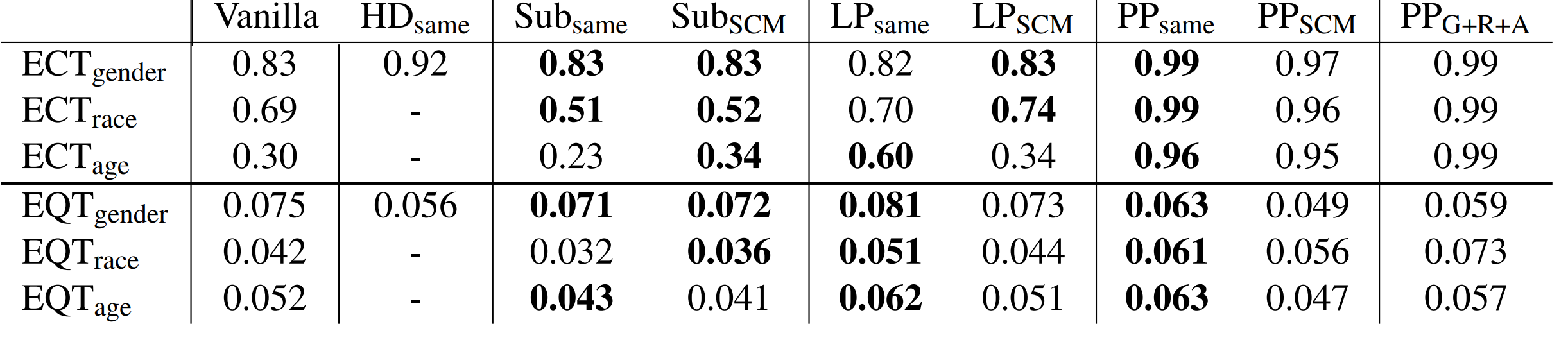}
  \captionof{table}{ECT and EQT scores reported by Omrani et al. for single-dimensional bias mitigation. These results were produced using the Word2Vec word embedding model.}
  \label{tab:Omrani}
\end{center}

There are limits to keep in mind. SCM axes reflect broad social judgments and may vary by culture \cite{ChenXia2023}. Compound contrastive lists were generated automatically for scale, which is efficient but less precise than hand curation. Finally, we did not measure downstream task impact, so the link from ECT and EQT to application performance remains indirect. Debiasing reduces specific, measurable projections but does not eliminate all forms of bias; hidden structure can persist in other directions \cite{Gonen2019}. The practical path forward is to combine SCM-based mitigation with task-level fairness checks and application-specific validation.

\section{Conclusions}
\label{sec:conclusion}

This section summarizes the key findings and insights obtained from the experiments on bias mitigation in aggregated word embeddings. These conclusions are drawn with consideration of the limitations discussed in Section \ref{sec:discussion}. Lastly, the contribution of this study to the existing research gap is outlined.

SCM-based bias mitigation, as performed by Omrani et al. on static word embeddings, can be extended from a single-dimensional to a multi-dimensional setting to address intersectional bias. This extension could be performed by aggregating word embeddings in two ways, namely, summing or concatenating the embeddings. As discussed in Section \ref{sec:results}, comparable results between both aggregation methods were observed. It was also noted that which aggregation method may perform better could be dependent on the word embedding model itself. These findings suggest that the methodology of Omrani et al. can be extended by aggregating word embeddings, which enables bias mitigation on intersectional biases.

Comparing compound word embeddings and singular word embeddings yielded mixed results. Bias mitigation in aggregated embeddings showed higher ECT scores and lower EQT scores than debiasing singular word embeddings. Interpreting these results requires some caution due to the critique that may be given to the chosen metrics for this study, as mentioned in Section \ref{sec:discussion}. Despite differences in ECT and EQT scores between bias mitigation in intersectional and singular biases, the values appear to be relatively close in both cases. Cautiously, it can be concluded that the quality and coherence of compound word embeddings are similar to those of singular word embeddings post-debiasing.

Next, the results across Word2Vec, GloVe, and Conceptnet Numberbatch can be compared. The results showed that the quality and coherence of the embeddings across different word embedding models seemed relatively consistent, with some minor differences discussed in Section \ref{sec:discussion}. Overall, it appears that the quality and coherence post-debiasing differ slightly across word embedding models.

To summarize these insights, the results are similar across aggregation methods, between singular and aggregated word embeddings, and across word embedding models. This indicates that extending SCM-based bias mitigation to address intersectional bias adequately preserves quality and coherence, making it a viable extension of bias mitigation beyond singular word embeddings.

Prior work on bias mitigation has largely been focused on singular word embeddings. This study explored the potential of extending an existing approach to address intersectional biases, thereby filling the research gap regarding the viability of SCM-based bias mitigation for intersectional biases.

\section*{Declarations}

\noindent \textbf{Data availability.}
All data used in this study are publicly available. We rely on pretrained static word embeddings (Word2Vec \cite{Mikolov2013}, GloVe \cite{Pennington2014}, and ConceptNet Numberbatch \cite{Speer2017}), the WikiText-103 corpus for light adaptation \cite{Merity2018}, SCM antonym lists derived from prior work \cite{Nicolas2021,Omrani2023}, and identity vocabularies adapted from \cite{Choenni2021}. Processed term lists and evaluation scripts sufficient to reproduce the reported ECT and EQT results are included with the submission or are available from the corresponding author on reasonable request.

\noindent \textbf{Competing interests.}
The authors declare that they have no competing interests, financial or non-financial.

\noindent \textbf{Funding.}
This research received no specific grant from any funding agency in the public, commercial, or not-for-profit sectors.

\noindent \textbf{Ethics approval and consent to participate.}
Not applicable. The study uses publicly available text resources and pretrained models and involves no human subjects or identifiable personal data.

\noindent \textbf{Consent for publication.}
Not applicable.

\noindent \textbf{Author contributions (CRediT).}
Eren Kocadag: Conceptualization; Methodology; Software; Validation; Formal analysis; Investigation; Data curation; Visualization; Writing, original draft; Writing, review and editing; Project administration.
Seyed Sahand Mohammadi Ziabari: Supervision; Conceptualization; Methodology; Writing, review and editing; Project administration; Corresponding author.
Ali Mohammed Mansoor Alsahag: Supervision; Methodology; Validation; Writing, review and editing.

All authors read and approved the final manuscript.

\noindent \textbf{Permissions for figures.}
All figures in this manuscript were created by the authors; no third-party images were used, and no permissions were required.

\bibliography{refs}

@inproceedings{Omrani2023,
  title     = {Social-group-agnostic bias mitigation via the stereotype content model},
  author    = {Omrani, Ali and Salkhordeh Ziabari, Alireza and Yu, Chenhao and Golazizian, Preni and Kennedy, Brendan and Atari, Mohammad and Ji, Heng and Dehghani, Morteza},
  booktitle = {Proceedings of the 61st Annual Meeting of the Association for Computational Linguistics},
  publisher = {Association for Computational Linguistics},
  year      = {2023},
  pages     = {4123--4139}
}

@incollection{Fiske2018,
  title     = {A model of (often mixed) stereotype content: Competence and warmth respectively follow from perceived status and competition},
  author    = {Fiske, Susan T. and Cuddy, Amy J. C. and Glick, Peter and Xu, Jun},
  booktitle = {Social Cognition},
  publisher = {Routledge},
  year      = {2018},
  pages     = {162--214}
}

@inproceedings{Papakyriakopoulos2020,
  title     = {Bias in word embeddings},
  author    = {Papakyriakopoulos, Orestis and Hegelich, Simon and Serrano, Juan Carlos Medina and Marco, Fabien},
  booktitle = {Proceedings of the 2020 Conference on Fairness, Accountability, and Transparency},
  publisher = {Association for Computing Machinery},
  year      = {2020},
  pages     = {446--457}
}

@article{Mikolov2013,
  title   = {Efficient estimation of word representations in vector space},
  author  = {Mikolov, Tomas},
  journal = {arXiv preprint arXiv:1301.3781},
  year    = {2013}
}

@inproceedings{Pennington2014,
  title     = {GloVe: Global vectors for word representation},
  author    = {Pennington, Jeffrey and Socher, Richard and Manning, Christopher D.},
  booktitle = {Proceedings of the 2014 Conference on Empirical Methods in Natural Language Processing},
  publisher = {Association for Computational Linguistics},
  year      = {2014},
  pages     = {1532--1543}
}

@article{Bengio2003,
  title   = {A neural probabilistic language model},
  author  = {Bengio, Yoshua and Ducharme, R{\'e}jean and Vincent, Pascal and Jauvin, Christian},
  journal = {Journal of Machine Learning Research},
  volume  = {3},
  pages   = {1137--1155},
  year    = {2003}
}

@article{Gupta2021,
  title   = {Obtaining better static word embeddings using contextual embedding models},
  author  = {Gupta, Pankaj and Jaggi, Martin},
  journal = {arXiv preprint arXiv:2106.04302},
  year    = {2021}
}

@article{Petreski2023,
  title   = {Word embeddings are biased. But whose bias are they reflecting?},
  author  = {Petreski, Damjan and Hashim, Ismail C.},
  journal = {AI \& Society},
  volume  = {38},
  number  = {2},
  pages   = {975--982},
  year    = {2023}
}

@inproceedings{Bolukbasi2016,
  title     = {Man is to computer programmer as woman is to homemaker? Debiasing word embeddings},
  author    = {Bolukbasi, Tolga and Chang, Kai-Wei and Zou, James Y. and Saligrama, Venkatesh and Kalai, Adam T.},
  booktitle = {Advances in Neural Information Processing Systems},
  year      = {2016}
}

@inproceedings{Manzini2019,
  title     = {Black is to criminal as Caucasian is to police: Detecting and removing multiclass bias in word embeddings},
  author    = {Manzini, Thomas and Lim, Yao Chong and Tsvetkov, Yulia and Black, Alan W.},
  booktitle = {Proceedings of the 2019 Conference of the North American Chapter of the Association for Computational Linguistics: Human Language Technologies},
  publisher = {Association for Computational Linguistics},
  year      = {2019}
}

@inproceedings{Dev2019,
  title     = {Attenuating bias in word vectors},
  author    = {Dev, Sunipa and Phillips, Jeff},
  booktitle = {Proceedings of the Twenty-Second International Conference on Artificial Intelligence and Statistics},
  series    = {Proceedings of Machine Learning Research},
  volume    = {89},
  pages     = {879--887},
  year      = {2019}
}

@article{Gonen2019,
  title   = {Lipstick on a pig: Debiasing methods cover up systematic gender biases in word embeddings but do not remove them},
  author  = {Gonen, Hila and Goldberg, Yoav},
  journal = {arXiv preprint arXiv:1903.03862},
  year    = {2019}
}

@inproceedings{Ghai2021,
  title     = {WordBias: An interactive visual tool for discovering intersectional biases encoded in word embeddings},
  author    = {Ghai, Bhavya and Hoque, Mohammed N. and Mueller, Klaus},
  booktitle = {Extended Abstracts of the 2021 CHI Conference on Human Factors in Computing Systems},
  year      = {2021},
  pages     = {1--7}
}

@article{Lepori2020,
  title   = {Unequal representations: Analyzing intersectional biases in word embeddings using representational similarity analysis},
  author  = {Lepori, Michael A.},
  journal = {arXiv preprint arXiv:2011.12086},
  year    = {2020}
}

@article{Friehs2022,
  title   = {Examining the structural validity of stereotype content scales},
  author  = {Friehs, Marie-Therese and Kotzur, Patrick F. and B{\"o}ttcher, Jan and Z{\"o}ller, Anne-Kathrin C. and L{\"u}ttmer, Tobias and Wagner, Ulrich and Asbrock, Frank and van Zalk, Maarten H. W.},
  journal = {International Review of Social Psychology},
  volume  = {35},
  number  = {1},
  year    = {2022}
}

@article{ChenXia2023,
  title   = {Cultural variations in perceptions and reactions to social norm transgressions},
  author  = {Chen-Xia, X. J. and Betancor, Ver{\'o}nica and Rodr{\'\i}guez-G{\'o}mez, Luis and Rodr{\'\i}guez-P{\'e}rez, Armando},
  journal = {Frontiers in Psychology},
  volume  = {14},
  year    = {2023}
}

@article{Nicolas2021,
  title   = {Comprehensive stereotype content dictionaries using a semi-automated method},
  author  = {Nicolas, Gilles and Bai, Xiaoyan and Fiske, Susan T.},
  journal = {European Journal of Social Psychology},
  volume  = {51},
  number  = {1},
  pages   = {178--196},
  year    = {2021}
}

@inproceedings{Choenni2021,
  title     = {Stepmothers are mean and academics are pretentious: What do pretrained language models learn about you?},
  author    = {Choenni, Rochelle and Shutova, Ekaterina and van Rooij, Robert},
  booktitle = {Proceedings of the 2021 Conference on Empirical Methods in Natural Language Processing},
  publisher = {Association for Computational Linguistics},
  year      = {2021},
  pages     = {1477--1491}
}

@inproceedings{Speer2017,
  title     = {ConceptNet 5.5: An open multilingual graph of general knowledge},
  author    = {Speer, Robyn and Chin, Joshua and Havasi, Catherine},
  booktitle = {Proceedings of the AAAI Conference on Artificial Intelligence},
  volume    = {31},
  number    = {1},
  year      = {2017}
}

@inproceedings{Merity2018,
  title     = {Scalable language modeling: WikiText-103 on a single GPU in 12 hours},
  author    = {Merity, Stephen and Keskar, Nitish Shirish and Bradbury, James and Socher, Richard},
  booktitle = {Proceedings of the SysML Conference},
  year      = {2018}
}

@inproceedings{Garbat2026WarmthCompetence,
  author    = {Garbat, A. and Mohammadi Ziabari, Seyed Sahand and Mansoor Alsahag, Ali Mohammed},
  title     = {Warmth and Competence Debiasing of Contextual Embeddings},
  booktitle = {Proceedings of the 20th IEEE International Conference on Semantic Computing},
  year      = {2026},
  note      = {Fill in pages/DOI/URL from the published version}
}

@inproceedings{Zhu2025TaskAdaptiveSCM,
  author    = {Zhu, C. and Mohammadi Ziabari, Seyed Sahand and Mansoor Alsahag, Ali Mohammed},
  title     = {Task-Adaptive Debiasing with {SCM} for Sentiment Analysis},
  booktitle = {Proceedings of [FILL THIS IN]},
  year      = {2025},
  note      = {Fill in venue/pages/DOI/URL once confirmed}
}

\end{document}